\documentclass[conference]{IEEEtran}

\usepackage{cite}
\usepackage{amsmath,amssymb,amsfonts}
\usepackage{algorithmic}
\usepackage{graphicx}
\usepackage{textcomp}
\usepackage{xcolor}

\usepackage{caption}
\usepackage{algorithm}
\usepackage{amsthm}
\usepackage{array}
\usepackage{balance}
\usepackage{booktabs}
\usepackage[edges]{forest}
\usepackage{makecell}
\usepackage{multicol}
\usepackage{multirow}
\usepackage{stfloats}
\usepackage{tabularx}
\usepackage{tikz}
\usepackage{url}
\usepackage{verbatim}
\usepackage{xcolor}
\usepackage{xspace}
\usepackage[hidelinks,
            colorlinks=true,
            linkcolor=blue,
            urlcolor=blue,
            citecolor=blue]{hyperref}

\definecolor{hidden-draw}{RGB}{20,68,106}
\definecolor{hidden-pink}{RGB}{255,245,247}
\definecolor{lightred}{RGB}{255, 204, 204}
\definecolor{lightgreen}{RGB}{224, 255, 225}
\definecolor{lightyellow}{RGB}{255, 241, 224}
\definecolor{lightpurple}{RGB}{225, 225, 255}
\definecolor{lightgray}{gray}{0.9}
\definecolor{text-red}{RGB}{255, 0, 0}
\definecolor{text-blue}{RGB}{0, 0, 255}
\definecolor{deep-purple}{RGB}{84, 74, 255}
\definecolor{deep-blue}{RGB}{0, 170, 238}
\definecolor{deep-green}{RGB}{63, 183, 4}

\newcommand{\txrd}[1]{\textcolor{text-red}{\textbf{#1}}}
\newcommand{\txbl}[1]{\textcolor{text-blue}{\textbf{#1}}}

\newcommand{\xs}[1]{\textcolor{text-red}{#1}}

\def\model{\textbf{UniMamba}}

\renewcommand{\txrd}[1]{\textbf{#1}}
\renewcommand{\txbl}[1]{\underline{#1}}
\renewcommand{\xs}[1]{\textcolor{black}{#1}}

\def\BibTeX{{\rm B\kern-.05em{\sc i\kern-.025em b}\kern-.08em
    T\kern-.1667em\lower.7ex\hbox{E}\kern-.125emX}}
	
\begin{document}

\title{UniMamba: A Unified Spatial-Temporal Modeling Framework with State-Space and Attention Integration}

\author{
	\IEEEauthorblockN{Xingsheng Chen\textsuperscript{1}, Xianpei Mu\textsuperscript{2}, Deyu Yi\textsuperscript{3}, Yilin Yuan\textsuperscript{2}, Xingwei He\textsuperscript{1},}
	\IEEEauthorblockN{Bo Gao\textsuperscript{2}, Regina Zhang\textsuperscript{4}, Pietro Lio\textsuperscript{4}, Siu-Ming Yiu\textsuperscript{1}}
	\IEEEauthorblockA{\textsuperscript{1}School of Computing and Data Science, The University of Hong Kong, Hong Kong, China}
	\IEEEauthorblockA{\textsuperscript{2}School of Information Engineering, Beijing Institute of Graphic Communication, Beijing, China}
	\IEEEauthorblockA{\textsuperscript{3}Innovation Engineering College, Macau University of Science and Technology, Macau, China}
	\IEEEauthorblockA{\textsuperscript{4}Department of Computer Science and Technology, University of Cambridge, Cambridge, UK}
}

\maketitle

\begin{abstract}
	Multivariate time series forecasting is fundamental to numerous domains such as energy, finance, and environmental monitoring, where complex temporal dependencies and cross-variable interactions pose enduring challenges. Existing Transformer-based methods capture temporal correlations through attention mechanisms but suffer from quadratic computational cost, while state-space models like Mamba achieve efficient long-context modeling yet lack explicit temporal pattern recognition. Therefore we introduce \textbf{\model}, a unified spatial-temporal forecasting framework that integrates efficient state-space dynamics with attention-based dependency learning. \model\ employs a \textbf{Mamba Variate–Channel Encoding Layer} enhanced with \xs{FFT-Laplace Transform} and TCN to capture global temporal dependencies, and a \textbf{Spatial Temporal Attention Layer} to jointly model inter-variate correlations and temporal evolution. A \textbf{Feedforward Temporal Dynamics Layer} further fuses continuous and discrete contexts for accurate forecasting. Comprehensive experiments on \xs{eight} public benchmark datasets demonstrate that \model\ consistently outperforms state-of-the-art forecasting models in both forecasting accuracy and computational efficiency, establishing a scalable and robust solution for long-sequence multivariate time-series prediction. The code is available at~\url{https://github.com/XsChen524/unimamba-ts}
\end{abstract}

\section{Introduction}

Time-series forecasting plays a critical role in a wide range of real-world applications, including energy demand prediction \cite{muralitharan2018neural}, financial forecasting \cite{xing2018natural}, traffic management \cite{zhang2025efficient,zhang2024survey}, and environmental monitoring \cite{kumar2012environmental}. The task remains inherently challenging due to complex temporal dependencies \cite{chen2026mode}, multi-variate correlations \cite{liu2023itransformer}, and non-stationary dynamics \cite{zhang2025autohformer} commonly observed in practical data. As forecasting horizons grow longer and data modalities become more heterogeneous, building models that can efficiently capture both \emph{global temporal patterns} and \emph{cross-variate relationships} has become increasingly vital.

Early approaches relied on statistical and recurrent models such as ARIMA \cite{shumway2017arima}, LSTM \cite{greff2016lstm}, and GRU \cite{dey2017gate}, which primarily captured short-term dependencies but struggled with scalability and stability over extended sequences. The advent of the Transformer \cite{vaswani2017attention} introduced a new era for sequence modeling by leveraging self-attention to capture long-range interactions. Transformer-based forecasting frameworks \cite{lim2021time,liu2021pyraformer,zhang2022crossformer} have demonstrated impressive accuracy; however, their quadratic time complexity and high memory cost severely limit scalability for long and high-dimensional time series. Attempts to linearize attention \cite{zeng2023transformers,zhou2022fedformer} mitigate efficiency issues, but often lead to degraded temporal precision and reduced robustness.

Recently, state-space models (SSMs) \cite{merrill2024illusion} have reemerged as an efficient and theoretically grounded alternative. The Mamba architecture \cite{gu2023mamba} and its variants \cite{zhang2025fldmamba} introduce selective scanning mechanisms and continuous-time dynamics, offering superior scalability in handling long sequences. Despite their success, existing Mamba-based frameworks \cite{gu2023mamba,wang2024mamba,liang2024bi} mainly focus on one-dimensional temporal dynamics and overlook explicit spatial or cross-variate dependencies \cite{liu2023itransformer,li2023revisiting}, which are essential in multivariate forecasting \cite{patro2024simba}.

To bridge these two paradigms, we propose \textbf{\model}, a unified spatial-temporal forecasting framework that combines frequency analysis and dependencies capture in bidirectional Mamba variant with expressive power of attention mechanisms. \model\ consists of a \textbf{Mamba Variate–Channel (VC) Encoding Layer} enhanced with Fast Fourier Transform(FFT)-Laplace reconstruction and Temporal Convolution Networks (TCN) for dynamic feature propagation, followed by a \textbf{spatial temporal attention Layer} to jointly model inter-variates and temporal dependencies. A \textbf{Feedforward Temporal Dynamics (FFN TD) Layer} adaptively fuses global and local temporal contexts before projection, enabling robust long-horizon prediction.

Our primary contributions can be summarized as follows:
\begin{itemize}
    \item We propose \textbf{\model}, the first unified framework that combines state-space dynamics with spatial temporal attention, effectively leveraging the strengths of both Transformers and Mamba architectures.
    \item We design an \textbf{enhanced Mamba Variate–Channel Encoding Layer} incorporating FFT-Laplace transform and TCN modules, enabling efficient modeling of complex interdependencies across time and variables.
    \item We demonstrate that \model\ achieves state-of-the-art forecasting performance across multiple public datasets, outperforming Transformer, MLP and Mamba-based baselines in terms of accuracy, scalability, and robustness.
\end{itemize}

By coupling attention-driven spatial temporal modeling with efficient state space recurrent dynamics, \model\ provides a powerful yet scalable solution for long-horizon multivariate time-series forecasting for production.

\section{Related Work}

\subsection{Transformer-Based Spatial Temporal Forecasting}
Transformer-based architectures \cite{vaswani2017attention} have substantially influenced time-series forecasting by enabling long-range temporal dependency modeling through self-attention. Subsequent works have refined this paradigm for temporal tasks, such as employing causal masking \cite{lim2021time,torres2021deep,zheng2021graph} to maintain sequence order or using multi-scale hierarchical designs like Pyraformer \cite{liu2021pyraformer}. Moreover, cross-variate modules such as Crossformer \cite{zhang2022crossformer} enhance inter-variable interactions through multidimensional attention. Despite these improvements, Transformer variants still suffer from quadratic computational cost $\mathcal{O}(L^2)$ and substantial memory overhead, which restrict scalability for long or multivariate sequences. To alleviate this, recent studies explore linearized \cite{zeng2023transformers,zhou2022fedformer} or patch-based \cite{huang2024long} mechanisms, but these often limit global context modeling.  

\subsection{State-Space and Mamba-Based Models}
State space models (SSMs) have recently offered an efficient alternative for long-horizon sequence modeling. The Mamba SSM \cite{gu2023mamba} introduces a selective scanning mechanism to capture temporal dynamics in linear time, significantly improving training efficiency and scalability over attention-based models. Variants such as S-Mamba \cite{wang2024mamba} fuses \xs{bidirectional Mamba branches that processing forward and flipped sequences} to strengthen global pattern recognition. However, existing Mamba architectures primarily focus on propagating or discarding information in sequences and maintaining hidden states precisely, offering limited capacity to decompose and reconstruct temporal signals. These are critical gaps for real-world multivariate forecasting tasks.

\subsection{Our Contributions: The \model\ Framework}
To overcome these limitations, we propose \textbf{\model}, a unified hybrid framework that seamlessly integrates enhanced Mamba Variate–Channel encoding and Spatial Temporal Attention. \model\ consist of following components to model temporal signals, capture dependencies and generate precies forecasting results.
\textbf{1. Enhanced Mamba Variate–Channel Encoding:} We incorporate Mamba-based selective scanning with \xs{FFT and learnable Laplace inverse transform for temporal signal reconstruction and TCN for local smoothing} to build a more expressive and stable sequence representation.
\textbf{2. Spatial Temporal Attention:} Unlike conventional attention models that treat variables and time independently, \model\ introduces a joint spatial temporal attention layer to strengthen its recognition and expressiveness across both dimensions.
\textbf{3. Temporal–Dynamic Feedforward Layer:} The FFN TD Layer further refines forecasting results by adaptively normalization and feed forwarding tensors before final projection.

Through this unified design, \textbf{\model} combines advantages of both attention and state-space architectures, offering a highly efficient, scalable, and robust framework for multivariate time-series forecasting.

\begin{figure*}[!t]
    \centering
    \includegraphics[width=\linewidth]{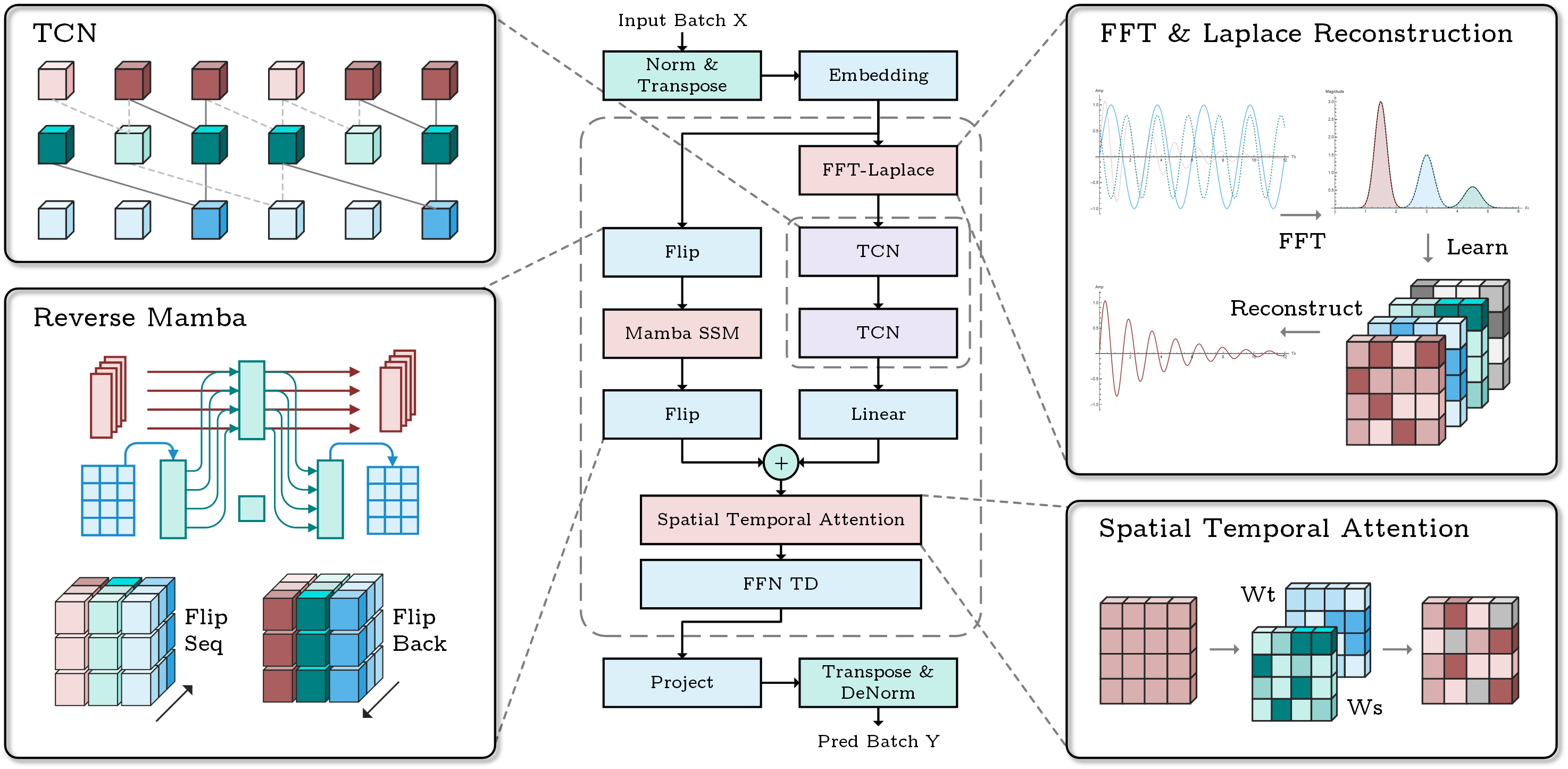}
    \caption{Framework of~\model}
    \label{fig:framework}
\end{figure*}
\section{Methodology}

\subsection{Problem Formulation}

Let $\mathbf{X} = [\mathbf{x}_1, \mathbf{x}_2, \ldots, \mathbf{x}_T] \in \mathbb{R}^{T \times D}$ denote a multivariate time series with $T$ time steps and $D$ variables. Each vector $\mathbf{x}_t = [x_t^{(1)}, x_t^{(2)}, \ldots, x_t^{(D)}]$ represents the observations of $D$ correlated signals at time step $t$. Given a historical observation window of length $L$, the forecasting objective is to predict the next $H$ time steps:
\begin{equation}
    \hat{\mathbf{Y}} = [\hat{\mathbf{x}}_{T+1}, \hat{\mathbf{x}}_{T+2}, \ldots, \hat{\mathbf{x}}_{T+H}] = f_\theta(\mathbf{X}_{1:L})
\end{equation}
where $f_\theta(\cdot)$ represents a learnable forecasting model parameterized by $\theta$, and $\hat{\mathbf{Y}}$ is the model’s predicted future sequence.

In multivariate forecasting, two major challenges must be addressed:  
(\textit{i}) capturing long-range temporal dependencies that span wide contextual horizons, and  
(\textit{ii}) modeling spatial (cross-variate) relationships among multiple correlated variables.  
Conventional recurrent or attention-based models often treat these dependencies separately or require quadratic computation, leading to inefficiency and limited scalability, especially when $L$ and $D$ become large.

\subsection{Overall Architecture}

\textbf{\model} introduces a unified architecture for efficient and robust multivariate time-series forecasting. As illustrated in Figure~\ref{fig:framework}, \xs{input batch is normalized and got its variate and time sequence dimension transposed before embedding, because tokenization along sequence dimension improves model's understanding quality~\cite{liu2023itransformer}. Embedded input sequence are fed into bidirectional branches. The embedded sequence in forward branch are transformed to frequency domain via FFT to filter out aperiodic signals and disturbance. Subsequently temporal signal is reconstructed by learnable Laplace transform to represent exponential trends, transients and periodicity across variate channels. TCNs then extract localized and medium-range dependencies, after which tensors are fed to MLP to match dimensions for fusion. While the reserve Mamba branch encodes long-term temporal evolution in backward direction to capture dependencies in future time steps which are neglected by causal convolution}. The spatial temporal attention layer adaptively adjusts weight matrices to capture variates and temporal sequence interactions across time. Finally FFN TD block refines latent embeddings before the tensor being projected and de-normalized to output domain.
\xs{\begin{equation}
\begin{gathered}
    H = \big(\text{Tokenize}(\text{Transpose}(\mathbf{X}_{1:L}))\big) \\
    \hat{\mathbf{Y}} = \text{Proj}\big(\text{FFN-TD}(\text{ST-Attn}(FW(H) + BW(H)))\big)
\end{gathered}
\end{equation}}
Formally, the overall forecasting pipeline of \model\ is expressed as above. This unified formulation enables \model\ to jointly \xs{reconstruct trend and seasonal patterns across temporal signal channels in training, inspect dependencies of various scales, and model spatial and temporal correlations} while maintaining computational efficiency and generality in real world scenarios.

\subsection{Learnable Frequency Domain Encoding}

To capture both transients and long-range seasonal patterns, \model\ employs an FFT and learnable Laplace reconstruction as shown in Algorithm~\ref{alg:fft-laplace}. The model performs FFT on $D$-length sequence in each channel and transform signals to frequency domain through $H_{v,k} = \sum_{d} x_{v,d}e^{\frac{-2\pi i k d}{D}}, d, k \in \{1,2, \dots, D\}$, where $v$ and $d$ are indices of channels and samples in temporal signals. $H_{v,k}$ is $k$-th frequency-domain components in $v$-th channel. To capture long-term seasonal patterns in temporal signals and properly handle negative impacts of transient patterns, the model adopts Laplace signal reconstruction implemented with neural network components:
\begin{equation}
\hat{\mathbf{Y}} = \text{Laplace}(H) = \sum_{v=1}^{V} A \cdot e^{\alpha t} \cdot \cos(\omega t + \phi)
\end{equation}
where $A$, $\alpha$, $\omega$, and $\phi$ are learnable reconstruction parameters trained with input tensors. $e^{\alpha t}$ term contributes to modeling transient dynamics and $\cos(\omega t + \phi)$ recovers complex synthesis of periodic signals. To a satisfactory confidence, temporal signals with instantaneous and periodic patterns are reconstructed across given contextual ranges. \xs{such signal reconstruction enhances the representation of non-stationary dynamics in discrete temporal sequence. Neural operators are computed from projections during training process with optional low-rank approximation to adjust parameter size level.}

\begin{algorithm}[tb]
    \caption{Learnable FFT and Laplace Reconstruction}
	\label{alg:fft-laplace}
	\begin{algorithmic}
		\REQUIRE \textbf{Input} $X \in \mathbb{R}^{B \times V \times D}$
		\STATE $H \leftarrow \text{FFT}(X)$, $X_{real} \leftarrow \text{Re}(H)$, $X_{imag} \leftarrow \text{Im}(H)$
        \IF{$\text{Low Rank}$}
    		\STATE $A = U_A \in [B,V,P,R] \times V_A^{\textsf{T}}  \leftarrow \text{Projector}_A(X)$
        \ELSE
            \STATE $A \in [B,V,P,P] \leftarrow \text{Projector}_A(X)$
        \ENDIF
        \STATE $\alpha \leftarrow \text{Projector}_\alpha(X_{real})$, $\omega,\phi \leftarrow \text{Projector}_{\omega,\phi}(X_{imag})$
		\STATE $t: [T] \leftarrow \text{Projector}_t(\text{linspace}(0.0001, 1, T))$
		\STATE $\alpha \leftarrow -\text{ELU}(-\alpha)$ [Ensure decay]
		\STATE $\hat{Y}:[B,V,P] \leftarrow \sum_h A \cdot \exp(\alpha \cdot t) \cdot \cos(\omega \cdot t + \phi)$
		\STATE \textbf{Output} $\hat{Y} \in \mathbb{R}^{B \times V \times P}$
	\end{algorithmic}
\end{algorithm}


\subsection{Temporal Convolutional Networks}
The TCN layer captures local continuity and mid-term dependencies through dilated causal convolutions. The output of $l$-th TCN layer is as follow, where $W^{l}$ is $l$-th conv kernel, $d=2^{\,l-1}$ is dilation rate, and $\sigma$ is activation function:
\begin{equation}
    H^{(l)} = \sigma \bigl(W^{(l)} \ast_{d=2^{\,l-1}} H^{(l-1)}\bigr), H^{(0)} \in \mathbb{R}^{B \times V \times P}
\end{equation}
TCN layers process reconstructed temporal signals, expanding receptive fields for capturing short and mid-term dependencies and stabilize gradient propagation during long-sequence forecasting, without significantly increasing the model parameter size.

\subsection{Mamba State–Space Module (SSM) and Fusion}

\xs{On the reverse branch, \textit{Mamba} block scans flipped-over series and models temporal evolution backwards to comprehensively consider future information neglected by causal convolutions in forward branch. As in Algorithm~\ref{alg:unimamba}, the input sequence is flipped to obtain backward flow, ensuring global temporal information is considered with selective scan mechanism of linear complexity. After flipped back, the results of flipped \textit{Mamba} is linearly fused with forward branch and residual block, providing a complete temporal context:} $H_{l} = H_{l-1} + H_{FW} + H_{BW(Mamba)}$. This architecture combines the recurrence efficiency of state-space models \xs{with high quality signal reconstructions and broad receptive fields.}

\subsection{Spatial Temporal Attention Integration}

The Spatial Temporal Attention module adaptively captures variable-wise and temporal correlations by computing attention scores across both dimensions. As shown below, $H \in \mathbb{R}^{B \times V \times D}$ is fused input, $W$ is weight matrix, $V$ is contextual vector, and $\alpha$ is attention score matrix:
\begin{equation}
\begin{gathered}
    E_t=\text{ReLU}(W_tH+b_t), E_s = \text{ReLU}(W_s H^T + b_s) \\
    \alpha = \frac{\exp(V^T E)}{\sum_{j=1}^{N} \exp(V^T E[:, j])}
\end{gathered}
\end{equation}
Thus the reweighed output is $H_{l} = \frac{1}{2}(H_{t} + H_{s})$. It adjusts contextual embeddings dynamically, allowing the model to prioritize information-rich segments and mitigate noise or redundancy in multivariate inputs.

\subsection{Feedforward Temporal Dynamics and Projection}

A Feedforward Temporal Dynamics (FFN–TD) layer refines the fused representations by modeling residual dependencies and enhancing temporal smoothness. After passing all encoders, the final \xs{Projection, Transpose and DeNorm} layers project and reshape tensor of features into the target dimensionality, generating the forecasting output $\hat{\mathbf{Y}}$.

\subsection{Model Complexity Analysis}

\xs{\model's time complexity originates from projector's matrix computation and selective scan in two branches, and spatial temporal attention module. When $X \in \mathbb{R}^{T \times D}$ is fed into $D$-dimension model with low-rank $r$ for forecasting $P$ steps, the complexity is $\mathcal{O}(N \cdot P^2)$ (or $\mathcal{O}(N \cdot P \cdot r), r \ll P$ with approximation) for signal reconstruction, $O(N \cdot P)$ for TCNs, and $O(N \cdot D)$ for \textit{Mamba} block. Spatial temporal attention requires $O(N,D,E)$ time where $E$ is the attention dimension. Unlike Transformer-based models with self attention, UniMamba projects $L$ into model dimension and has its complexity dominated by the controllable parameter $D$, thereby avoiding the quadratic dependency $\mathcal{O}(L^2)$. The low-rank $r$ also helps maintain linear complexity given increasing output length $P$ in long-term forecasting task.} All above makes \model\ particularly well-suited for long-term forecasting and real-time inference on multivariate data streams.

\begin{algorithm}[tb]
    \caption{Forecast with UniMamba Encoders}
	\label{alg:unimamba}
	\begin{algorithmic}
		\REQUIRE \textbf{Input Batch} $X \in \mathbb{R}^{B \times L \times V}$
        \STATE $X_{in}^\top: [B, V, L] \leftarrow \text{Transpose}(\text{Normalize}(X))$
        \STATE $H_0: [B, V, D] \leftarrow \text{Embedding}(\text{Time Feature}(X_{in}^\top))$
		\FOR{$l = 1$ to $E$}
		\STATE \textbf{Forward}: $H_{fft}: [B, V, L] \leftarrow \text{FFT-Laplace}(H_{l-1})$
		\FOR{$k = 1$ to $K$}
		\STATE $H_{fft}: [B, V, L] \leftarrow \text{TCN}_{k} (H_{fft})$
		\ENDFOR
		\STATE $H_{FW}: [B, V, D] \leftarrow \text{Linear}(H_{fft})$
		\STATE \textbf{Backward}: $H_{flip}: [B, V, D] \leftarrow \text{flip}(H_{l-1})$
		\STATE $H_{Mamba}: [B, V, D] \leftarrow \text{MambaSSM}((H_{flip}))$
		\STATE $H_{BW}: [B, V, D] \leftarrow \text{flip}(H_{Mamba})$
		\STATE \textbf{Fusion}: $H_{l}: [B, V, D] \leftarrow H_{l-1} + H_{FW} + H_{BW}$
		\STATE $H_{l}: [B, V, D] \leftarrow \text{Spatial Temporal Attn}(H_{l-1})$
		\STATE $H_{l}: [B, V, D] \leftarrow \text{Norm}\big(H_{l-1} + \text{FFN}(\text{Norm}(H_{l-1}))\big)$
		\ENDFOR
		\STATE $\hat{Y}: [B, P, V] \leftarrow \text{DeNorm}(\text{Linear}(H_E)[:, :V, :]^\top)$
		\STATE \textbf{Output Batch} $\hat{Y} \in \mathbb{R}^{B \times P \times V}$
	\end{algorithmic}
\end{algorithm}


\section{Experiments}

We conduct experiments to answer the following research questions: \textbf{1)} How does \model~matches or outperforms current outstanding baselines methods in terms of overall performance.
\textbf{2)} What role does each individual component of \model\ play in contributing to overall effectiveness. \textbf{3)} In terms of computational efficiency, how does \model\ compare with leading forecasting models. \textbf{4)} \xs{To what extent \model\ maintains robustness under noise insertion.} \textbf{5)} How does altering the lookback window length influence \model’s long-range forecasting accuracy relative to other models? \textbf{6)} How effectively can \model\ identify and represent transient behaviors and temporal structures, including challenging or edge-case scenarios and known limitations?

\subsection{Experimental Setup}

\textbf{Datasets:} We evaluate our approach using eight publicly available benchmark datasets, namely ETTh1, ETTh2, ETTm1, ETTm2, Exchange, Weather, Solar-Energy, and PEMS08, as summarized in Table~\ref{tab:dataset_overview}. These datasets span multiple application domains and exhibit diverse temporal and structural properties. MSE and MAE are used as evaluation metrics.

\textbf{Baselines:} We compare \model\ with \xs{nine leading forecasting models} that collectively represent three primary architectural families: Transformer-based (six methods), MLP-based (two methods), and state-space model (SSM)-based (one method). The Transformer family includes Autoformer \cite{wu2021autoformer}, which combines time-series decomposition with an autocorrelation mechanism to capture periodic patterns without relying on standard self-attention; FEDformer \cite{zhou2022fedformer}, which replaces standard attention operations with frequency-domain representations to achieve greater efficiency while retaining a broad receptive field; and Crossformer \cite{zhang2022crossformer}, which adopts multi-dimensional attention applied to patched subsequences, enhancing local feature learning though its performance may taper off for very long horizons. DLinear \cite{zeng2023transformers} demonstrates that a pair of lightweight linear layers operating on decomposed trend and residual series can rival more intricate attention models across diverse tasks. PatchTST \cite{huang2024long} relies on segmented, channel-wise embeddings to extract temporal cues across multiple scales, while iTransformer \cite{liu2023itransformer} inverses the standard attention layout to emphasize relationships among variables, though its flat MLP-based tokenization struggles to represent hierarchical time dependencies.  
The MLP-based group consists of TimesNet \cite{wu2022timesnet}, which maps one-dimensional sequences into two-dimensional periodic tensors to jointly learn intra- and inter-period patterns. And TiDE \cite{das2023long}, which structures stacked fully connected layers in an encoder–decoder arrangement, discarding both attention and recurrence while maintaining strong temporal modeling capacity.  
Lastly, the SSM-based model S-Mamba \cite{wang2025mamba} utilizes per-variable tokenization combined with bidirectional Mamba modules to represent variable interactions, further strengthened by feed-forward layers that capture temporal transitions.


\begin{table}[htbp]
\vspace{-0.1in}
    \centering
    \caption{Overview of 8 publicly time-series datasets.}\vspace{-0.1in}
    \label{tab:dataset_overview}
    \resizebox{0.98\linewidth}{!}{
    \renewcommand{\arraystretch}{0.85}
    \begin{tabular}{lccc}
        \toprule
        \textbf{Dataset} & \textbf{Variables} & \textbf{Total Time Steps} & \textbf{Sampling Interval} \\
        \midrule
        ETTh1 & 7 & 17,420 & 1 hour \\
        ETTh2 & 7 & 17,420 & 1 hour \\
        ETTm1 & 7 & 69,680 & 15 minutes \\
        ETTm2 & 7 & 69,680 & 15 minutes \\
        Exchange & 8 & 7,588 & 1 day \\
        Weather & 21 & 52,696 & 10 minutes \\
        Solar Energy & 137 & 52,560 & 1 hour \\
        PEMS08 & 170 & 17,856 & 5 minutes \\
        \bottomrule
    \end{tabular}}
\end{table}

\subsection{Effectiveness}

We conduct effectiveness experiments on \model~with input sequence $L=96$ and forecast horizons $T \in \{96, 192, 336, 720\}$. Table~\ref{tab:effectiveness} presents comparisons across four datasets. \model\ consistently secures the best or second-best MSE and MAE scores for most forecasting horizons on given datasets, demonstrating superior effectiveness compared to state-of-the-art baselines such as S-Mamba and iTransformer. \xs{Comprehensive effectiveness experimental results and optimal hyperparameter settings are listed in appendix.}

The superior performance of~\textbf{\model} can be traced to its unified spatial temporal modeling design. 
First, \xs{the incorporation of FFT and learnable Laplace reconstruction modules enable accurate signal reconstruction and spectral representation, capturing periodic frequency-domain patterns and divergent trend without huge overheads. Second, TCN stack enhances local contextual modeling by capturing short-term temporal dependencies missed by purely spectral methods. 
Third, Mamba block models sequential dynamics in backward directions, enabling the framework to learn temporal correlations in future time steps ignored by causal convolutions.} Additionally, spatial temporal attention module adaptively integrates spatial correlations and temporal dependencies, allowing the model to focus on dynamic inter-variable relationships even in non-stationary settings. FFN TD and the final projection layers ensure stable feature refinement and precise mapping to prediction space.

These design choices collectively balance global and local temporal dynamics, enhance generalization across diverse domains, \xs{without involving enormous computational costs.} This explains \model ’s excellent and consistent prediction accuracy across different datasets and forecast horizons.


\begin{table*}[htb!]
	\vspace{-0.1in}
	\caption{
		Comparison results between \model\ and baselines on ETTm2, ETTh2, Weather, PEMS08 datasets in effectiveness experiments. \textbf{Bold} font denotes the best model and \underline{underline} denotes the second best. All baseline results are obtained from \cite{wang2025mamba}.}
	\label{tab:effectiveness}
	\renewcommand{\arraystretch}{1.0}
	\centering
	\resizebox{\textwidth}{!}{
		\begin{small}
			\setlength{\tabcolsep}{2.6pt}
			\vspace{1mm}
			\begin{tabular}{c|c|cc|cc|cc|cc|cc|cc|cc|cc|cc|cc}
				\toprule
				\multicolumn{2}{c|}{Models}
					& \multicolumn{2}{c|}{\textbf{\model}}
					& \multicolumn{2}{c|}{S-Mamba}
					& \multicolumn{2}{c|}{iTransformer}
					& \multicolumn{2}{c|}{PatchTST}
					& \multicolumn{2}{c|}{Crossformer}
					& \multicolumn{2}{c|}{TiDE}
					& \multicolumn{2}{c|}{TimesNet}
					& \multicolumn{2}{c|}{DLinear}
					& \multicolumn{2}{c|}{FEDformer}
					& \multicolumn{2}{c}{Autoformer}
				\\

				\cmidrule(lr){1-2}
				\cmidrule(lr){3-4}
				\cmidrule(lr){5-6}
				\cmidrule(lr){7-8}
				\cmidrule(lr){9-10}
				\cmidrule(lr){11-12}
				\cmidrule(lr){13-14}
				\cmidrule(lr){15-16}
				\cmidrule(lr){17-18}
				\cmidrule(lr){19-20}
				\cmidrule(lr){21-22}

				\multicolumn{2}{c|}{Metric}
					& MSE & MAE
					& MSE & MAE
					& MSE & MAE
					& MSE & MAE
					& MSE & MAE
					& MSE & MAE
					& MSE & MAE
					& MSE & MAE
					& MSE & MAE
					& MSE & MAE
				\\
				\toprule
				
				\multirow{5}{*}{\rotatebox{90}{ETTm2}}
					& 96          & \txrd{0.174} & \txrd{0.257} 
					& 0.179       & 0.263         
					& 0.180       & 0.264         
					& \txbl{0.175}& \txbl{0.259}  
					& 0.287       & 0.366         
					& 0.207       & 0.305         
					& 0.187       & 0.267         
					& 0.193       & 0.292         
					& 0.203       & 0.287         
					& 0.255       & 0.339         
				\\

					& 192         & \txrd{0.240} & \txrd{0.302} 
					& 0.250       & 0.309       
					& 0.250       & 0.309       
					& \txbl{0.241}& \txrd{0.302}
					& 0.414       & 0.492       
					& 0.290       & 0.364       
					& 0.249       & 0.309       
					& 0.284       & 0.362       
					& 0.269       & 0.328       
					& 0.281       & 0.340       
				\\

					& 336         & \txrd{0.304} & \txrd{0.342} 
					& 0.312       & 0.349       
					& 0.311       & 0.348       
					& \txbl{0.305}& \txbl{0.343}
					& 0.597       & 0.542       
					& 0.377       & 0.422       
					& 0.321       & 0.351       
					& 0.369       & 0.427       
					& 0.325       & 0.366       
					& 0.339       & 0.372       
				\\

					& 720         & \txbl{0.403} & \txrd{0.400}
					& 0.411       & 0.406       
					& 0.412       & 0.407       
					& \txrd{0.402}& \txrd{0.400}
					& 1.730       & 1.042       
					& 0.558       & 0.524       
					& 0.408       & 0.403       
					& 0.554       & 0.522       
					& 0.421       & 0.415       
					& 0.433       & 0.432       
				\\

				\cmidrule(lr){2-22}
					& Avg         & \txrd{0.280} & \txrd{0.325}
					& 0.288       & 0.332       
					& 0.288       & 0.332       
					& \txbl{0.281}& \txbl{0.326}
					& 0.757       & 0.610       
					& 0.358       & 0.404       
					& 0.291       & 0.333       
					& 0.350       & 0.401       
					& 0.305       & 0.349       
					& 0.327       & 0.371       
				\\
				\midrule

				\multirow{5}{*}{\rotatebox{90}{ETTh2}}
					& 96          & \txrd{0.293} & \txrd{0.343} 
					& \txbl{0.296}& \txbl{0.348} 
					& 0.297       & 0.349        
					& 0.302       & \txbl{0.348} 
					& 0.745       & 0.584        
					& 0.400       & 0.440        
					& 0.340       & 0.374        
					& 0.333       & 0.387        
					& 0.358       & 0.397        
					& 0.346       & 0.388        
				\\

					& 192         & \txrd{0.371} & \txrd{0.394} 
					& \txbl{0.376}& \txbl{0.396} 
					& 0.380       & 0.400        
					& 0.388       & 0.400        
					& 0.877       & 0.656        
					& 0.528       & 0.509        
					& 0.402       & 0.414        
					& 0.477       & 0.476        
					& 0.429       & 0.439        
					& 0.456       & 0.452        
				\\

					& 336         & \txrd{0.416} & \txrd{0.429} 
					& \txbl{0.424}& \txbl{0.431} 
					& 0.428       & 0.432        
					& 0.426       & 0.433        
					& 1.043       & 0.731        
					& 0.643       & 0.571        
					& 0.452       & 0.452        
					& 0.594       & 0.541        
					& 0.496       & 0.487        
					& 0.482       & 0.486        
				\\

					& 720         & \txrd{0.411} & \txrd{0.434} 
					& \txbl{0.426}& \txbl{0.444} 
					& 0.427       & 0.445        
					& 0.431       & 0.446        
					& 1.104       & 0.763        
					& 0.874       & 0.679        
					& 0.462       & 0.468        
					& 0.831       & 0.657        
					& 0.463       & 0.474        
					& 0.515       & 0.511        
				\\

				\cmidrule(lr){2-22}
					& Avg         & \txrd{0.373} & \txrd{0.400} 
					& \txbl{0.381}& \txbl{0.405} 
					& 0.383       & 0.407        
					& 0.387       & 0.407        
					& 0.942       & 0.684        
					& 0.611       & 0.550        
					& 0.414       & 0.427        
					& 0.559       & 0.515        
					& 0.437       & 0.449        
					& 0.450       & 0.459        
				\\
				\midrule

				\multirow{5}{*}{\rotatebox{90}{Weather}}
					& 96          & \txrd{0.155} & \txrd{0.200} 
					& 0.169       & \txbl{0.210} 
					& 0.174       & 0.214        
					& 0.177       & 0.218        
					& \txbl{0.158} & 0.230       
					& 0.202       & 0.261        
					& 0.172       & 0.220        
					& 0.196       & 0.255        
					& 0.217       & 0.296        
					& 0.266       & 0.336        
				\\

					& 192         & \txbl{0.210} & \txrd{0.251} 
					& {0.214}     & \txbl{0.253} 
					& 0.221       & 0.254        
					& 0.225       & 0.259        
					& \txrd{0.206}& 0.277        
					& 0.242       & 0.298        
					& 0.219       & 0.261        
					& 0.237       & 0.296        
					& 0.276       & 0.336        
					& 0.307       & 0.367        
				\\

					& 336         & \txrd{0.268} & \txrd{0.294} 
					& {0.274}     & \txbl{0.296} 
					& 0.278       & \txbl{0.296} 
					& 0.278       & {0.297}      
					& \txbl{0.272}& 0.335        
					& 0.287       & 0.335        
					& 0.280       & 0.306        
					& 0.283       & 0.335        
					& 0.339       & 0.380        
					& 0.359       & 0.395        
				\\

					& 720         & \txbl{0.349} & \txrd{0.346} 
					& 0.353       & 0.348        
					& 0.358       & \txbl{0.347} 
					& 0.354       & 0.348        
					& 0.398       & 0.418        
					& {0.351}     & 0.386        
					& 0.365       & 0.359        
					& \txrd{0.345}& 0.381        
					& 0.403       & 0.428        
					& 0.419       & 0.428        
				\\

				\cmidrule(lr){2-22}
					& Avg         & \txrd{0.246} & \txrd{0.273} 
					& \txbl{0.253}& \txbl{0.277} 
					& 0.258       & 0.278        
					& 0.259       & 0.281        
					& 0.259       & 0.315        
					& 0.271       & 0.320        
					& 0.259       & 0.287        
					& 0.265       & 0.317        
					& 0.309       & 0.360        
					& 0.338       & 0.382        
				\\
				\midrule
                
				\multirow{5}{*}{\rotatebox{90}{PEMS08}}
					& 12          & \txrd{0.075} & \txrd{0.176} 
					& \txbl{0.076}& \txbl{0.178} 
					& 0.079       & 0.182        
					& 0.168       & 0.232        
					& 0.165       & 0.214        
					& 0.227       & 0.343        
					& 0.112       & 0.212        
					& 0.154       & 0.276        
					& 0.173       & 0.273        
					& 0.436       & 0.485        
				\\

					& 24          & \txrd{0.102} & \txrd{0.202} 
					& \txbl{0.104}& \txbl{0.209} 
					& {0.115}     & {0.219}      
					& 0.224       & 0.281        
					& 0.215       & 0.260        
					& 0.318       & 0.409        
					& 0.141       & 0.238        
					& 0.248       & 0.353        
					& 0.210       & 0.301        
					& 0.467       & 0.502        
				\\

					& 48          & \txrd{0.145} & \txbl{0.232} 
					& \txbl{0.167}& \txrd{0.228} 
					& {0.186}     & {0.235}      
					& 0.321       & 0.354        
					& 0.315       & 0.355        
					& 0.497       & 0.510        
					& 0.198       & 0.283        
					& 0.440       & 0.470        
					& 0.320       & 0.394        
					& 0.966       & 0.733        
				\\

					& 96          & \txrd{0.219}& \txbl{0.277} 
					& 0.245       & {0.280}     
					& \txbl{0.221}& \txrd{0.267}
					& 0.408       & 0.417       
					& 0.377       & 0.397       
					& 0.721       & 0.592       
					& 0.320       & 0.351       
					& 0.674       & 0.565       
					& 0.442       & 0.465       
					& 1.385       & 0.915       
				\\

				\cmidrule(lr){2-22}
					& Avg         &\txrd{0.135}& \txrd{0.222} 
					& \txbl{0.148}&\txbl{0.224}
					& {0.150}     & {0.226}    
					& 0.280       & 0.321      
					& 0.268       & 0.307      
					& 0.441       & 0.464      
					& 0.193       & 0.271      
					& 0.379       & 0.416      
					& 0.286       & 0.358      
					& 0.814       & 0.659      
				\\
				\bottomrule
			\end{tabular}
		\end{small}
	}
	\vspace{-0.1in}
\end{table*}


\subsection{Ablation Study}

Table~\ref{tab:ablation} presents the ablation study conducted on ETTm2, Weather, and PEMS08 datasets with consistent hyperparameter settings to investigate the contribution of each key component in the \textbf{\model} framework. The baseline corresponds to the complete \model\ model, while subsequent variants remove or modify critical modules such as FFT-Laplace reconstruction, TCN, and attention.

From the results, we observe that removing either the signal reconstruction or TCN modules degrades forecasting performance in most of forecasting horizons and datasets. In terms of ETTm2, removing FFT-Laplace increases the average MSE from \xs{\textbf{0.281} to \textbf{0.288}, indicating that the frequency-domain analysis and reconstruction provided by the block is essential for capturing periodic and trend patterns.} Similarly, eliminating TCN stack causes further degradation (average MSE rising from \textbf{0.246} to \textbf{0.252} on Weather), confirming the significance of local temporal pattern extraction.

Furthermore, replacing the combination of both signal reconstruction and TCN with MLP leads to huge accuracy drop, highlighting their strong complementarity. Spectral transformations enable long-term dependency modeling while convolutional filters strengthen short-term feature extraction. The attention-related variants also provide valuable insights: disabling Spatial Temporal Attention results in higher error rates, \xs{particularly on ETTm2 dataset,} underscoring its role in adaptively weighting diverse temporal and spatial interactions. Conversely, self attention provides excellent accuracy in PEMS08 short-term prediction. Such mechanism also improves stability for some longer horizons, proving beneficial in handling non-stationary and multi-variate dependencies with quadratic complexity.

The minimal version that retaining only reverse Mamba block exhibits the worst overall results, validating that \model ’s superior performance emerges from the synergy between its signal construction, TCN, and attention mechanisms. Collectively, these findings demonstrate that each component contributes uniquely to \model ’s capability in achieving accurate, robust, and efficient spatiotemporal prediction across diverse domains. 


\begin{table}[htb!]
	\vspace{-0.1in}
	\caption{
		Ablation study results for \model\ across ETTm2, Weather, and PEMS08 datasets. The baseline represents the complete \model\ architecture, while the others represent ones with critical components replaced or removed.
	}
	\label{tab:ablation}
	\renewcommand{\arraystretch}{1.0}
	\centering
	\begin{small}
		\centering
		\setlength{\tabcolsep}{3.2pt}
		\vspace{0mm}
		\scriptsize
		\begin{tabular}{l|c|cc|cc|cc}
			\toprule
			\rule{0pt}{4pt}
			\multirow{2}{*}{Model}
			& \multirow{2}{*}{Length}
			& \multicolumn{2}{c|}{\textbf{ETTm2}}
			& \multicolumn{2}{c|}{\textbf{Weather}}
			& \multicolumn{2}{c}{\textbf{PEMS08}}                                                                                             \\
			\cline{3-4} \cline{5-6} \cline{7-8}
			\rule{0pt}{8pt} & & MSE & MAE & MSE & MAE & MSE & MAE \\
			\midrule

			\multirow{4}{*}{\parbox{1.5cm}{Baseline}}
			& 96  & \txrd{0.174} & \txrd{0.257} & \txbl{0.157} & \txbl{0.202} & \txbl{0.076} & \txbl{0.176} \\
			& 192 & \txrd{0.241} & \txrd{0.303} & \txbl{0.210} & \txrd{0.250} & \txrd{0.102} & \txbl{0.202} \\
			& 336 & \txbl{0.304} & \txbl{0.342} & \txrd{0.269} & \txbl{0.295} & \txbl{0.145} & \txbl{0.236} \\
			& 720 & \txrd{0.403} & \txrd{0.400} & \txrd{0.350} & 0.348        & \txbl{0.225} & \txrd{0.278} \\
			\midrule

			\multirow{4}{*}{\parbox{1.5cm}{w/o                          \\FFT-Laplace}}
			& 96  & \txbl{0.177} & 0.261  & 0.161        & 0.205        & 0.081        & 0.184        \\
			& 192 & 0.250        & 0.311  & 0.212        & \txbl{0.251} & 0.115        & 0.221        \\
			& 336 & 0.312        & 0.349  & 0.273        & 0.297        & 0.161        & 0.251        \\
			& 720 & 0.413        & 0.406  & \txrd{0.350} & 0.347        & 0.275        & 0.315        \\
			\midrule

			\multirow{4}{*}{\parbox{1.5cm}{w/o                                                                                                                \\TCN}}
			& 96  & 0.179        & 0.263        & 0.166 & 0.209        & 0.085        & 0.189        \\
			& 192 & \txbl{0.244} & \txbl{0.305} & 0.215 & 0.254        & 0.124        & 0.228        \\
			& 336 & \txrd{0.302} & \txrd{0.341} & 0.273 & \txrd{0.294} & \txrd{0.140} & 0.245        \\
			& 720 & \txbl{0.405} & \txrd{0.400} & 0.353 & 0.348        & \txrd{0.221} & 0.304        \\
			\midrule

			\multirow{4}{*}{\parbox{1.5cm}{w/o                          \\FFT-Laplace\\\& TCN}}
			& 96  & 0.182  & 0.267 & 0.165 & 0.208 & 0.089 & 0.295 \\
			& 192 & 0.251  & 0.312 & 0.215 & 0.253 & 0.138 & 0.244 \\
			& 336 & 0.312  & 0.349 & 0.275 & 0.297 & 0.169 & 0.272 \\
			& 720 & 0.413  & 0.404 & 0.352 & 0.348 & 0.281 & 0.346 \\
			\midrule

			\multirow{4}{*}{\parbox{1.5cm}{Self                                                                                                               \\Attention}}
			& 96  & 0.179 & 0.262        & 0.160        & 0.207        & \txrd{0.074} & \txrd{0.173} \\
			& 192 & 0.245 & \txbl{0.305} & 0.212        & 0.253        & \txrd{0.101} & \txrd{0.199} \\
			& 336 & 0.310 & 0.345        & \txbl{0.270} & 0.297        & 1.070        & 0.804        \\
			& 720 & 0.430 & 0.414        & \txrd{0.350} & \txrd{0.345} & 0.234        & \txbl{0.281} \\
			\midrule

			\multirow{4}{*}{\parbox{1.5cm}{w/o                                                                                                                \\Attention}}
			& 96  & \txbl{0.177} & \txbl{0.260} & \txrd{0.155} & \txrd{0.200} & \txbl{0.076} & \txbl{0.176} \\
			& 192 & 0.246        & 0.306        & \txrd{0.209} & \txbl{0.251} & 0.104        & 0.204        \\
			& 336 & 0.305        & 0.343        & 0.271        & 0.297        & \txbl{0.145} & \txrd{0.235} \\
			& 720 & 0.428        & 0.411        & \txrd{0.350} & \txbl{0.346} & 0.242        & 0.287        \\
			\midrule

			\multirow{4}{*}{\parbox{1.5cm}{Minimal}}
			& 96  & 0.184   & 0.266  & 0.166  & 0.209        & 0.089        & 0.195        \\
			& 192 & 0.250   & 0.311  & 0.216  & 0.254        & 0.141        & 0.247        \\
			& 336 & 0.311   & 0.349  & 0.273  & \txbl{0.295} & 0.195        & 0.287        \\
			& 720 & 0.413   & 0.405  & 0.353  & 0.348        & 0.342        & 0.383        \\
			\bottomrule
		\end{tabular}
	\end{small}
	\vspace{-0.1in}
\end{table}


\subsection{Efficiency}
Table~\ref{tab:efficiency} compares the efficiency of \textbf{\model} with firstclass baselines on ETTm1 and Weather datasets in terms of prediction accuracy and training time. \model\ achieves the lowest or tied-lowest MSE and MAE while maintaining competitive training efficiency. On ETTm1, it matches PatchTST's accuracy but reduces training time by 72\%. \model\ surpasses given models with significant gains in precision on Weather and keep the training cost at an competitive low level. These results highlight \model ’s balanced trade-off between accuracy and computational cost, achieved through its lightweight state-space formulation, parallelizable frequency-domain modeling and transformation, and spatial temporal attention integration. In summary, \model\ demonstrates high effectiveness and scalability for real-time forecasting tasks.


\begin{table}[t]
	\centering
	\footnotesize
	\caption{Efficiency comparison between~\model~and other models, in terms of prediction errors and training time.}
	\label{tab:efficiency}
	\begin{small}
		\vspace{0mm}
		\scriptsize
		\setlength{\tabcolsep}{5pt}
		\begin{tabular}{c|ccccc}
			\toprule
			\multicolumn{6}{c}{\textbf{ETTm1}}\\
			\midrule
			Models        &\model       &S-Mamba &iTrans      & PatchTST      & AutoF  \\
			\midrule
			MSE           &\txrd{0.324} &0.341   &0.342       &\txrd{0.324}& 0.526  \\
			MAE           &\txrd{0.362} &0.371   &0.377       &\txrd{0.362}& 0.488  \\
			Training Time &39.63(ms/it) &25.02   &\txrd{14.54}& 141.12     & 45.79  \\
			Change (\%)   &-            &-36.9\% &-63.3\%     & +256.1\%   &+15.5\% \\
			\bottomrule
		\end{tabular}

		\vspace{4pt}

		\setlength{\tabcolsep}{5pt}
		\begin{tabular}{c|ccccc}
			\toprule
			\multicolumn{6}{c}{\textbf{Weather}}\\
			\midrule
			Models        & \model       & S-Mamba & iTrans      & PatchTST    & AutoF  \\
			\midrule
			MSE           & \txrd{0.156} & 0.168    & 0.176      & 0.183    & 0.323  \\
			MAE           & \txrd{0.200} & 0.211    & 0.215      & 0.222    & 0.373  \\
			Training Time & 41.49(ms/it) & 30.36    &\txrd{20.68}& 158.87   & 46.40  \\
			Change (\%)   & -            & -26.8\%  & -50.2\%    & +282.9\% & +11.8\%\\
			\bottomrule
		\end{tabular}
        
	\end{small}
\end{table}


\subsection{Robustness}

\begin{figure}[!t]
    \centering
    \includegraphics[width=\linewidth]{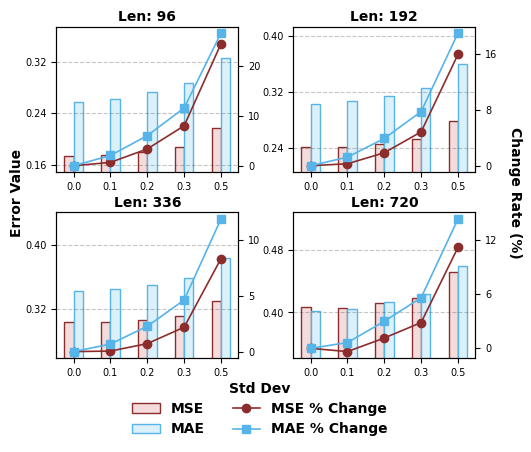}
    \caption{Robustness experiments on ETTm2}
    \label{fig:robustness}
\end{figure}

Figure~\ref{fig:robustness} shows the robustness analysis of \textbf{\model} on ETTm2 under increasing noise perturbations with forecast lengths $L \in \{96,192,336,720\}$. The figure reports MSE and MAE values with their percentage changes as standard deviation of gaussian noise grows.

Across all horizons, \model\ remains highly stable. When standard deviation of gaussian noise increases to 0.3, MSE increases by less than 10\%, indicating model's strong resistance to input distortion. Even at 0.5 noise's standard deviation, the error growth is moderate and does not exceed 20\% in long-term predictions. This robustness stems from the combined effect of Mamba SSM capturing global temporal patterns, Laplace Transform, which enhances frequency-domain stability and realistic signal reconstruction, and spatial temporal attention that adaptively adjusts feature weighting under noisy conditions.


\subsection{Lookback Length Study}
\begin{figure*}[!t]
    \centering
    \includegraphics[width=\linewidth]{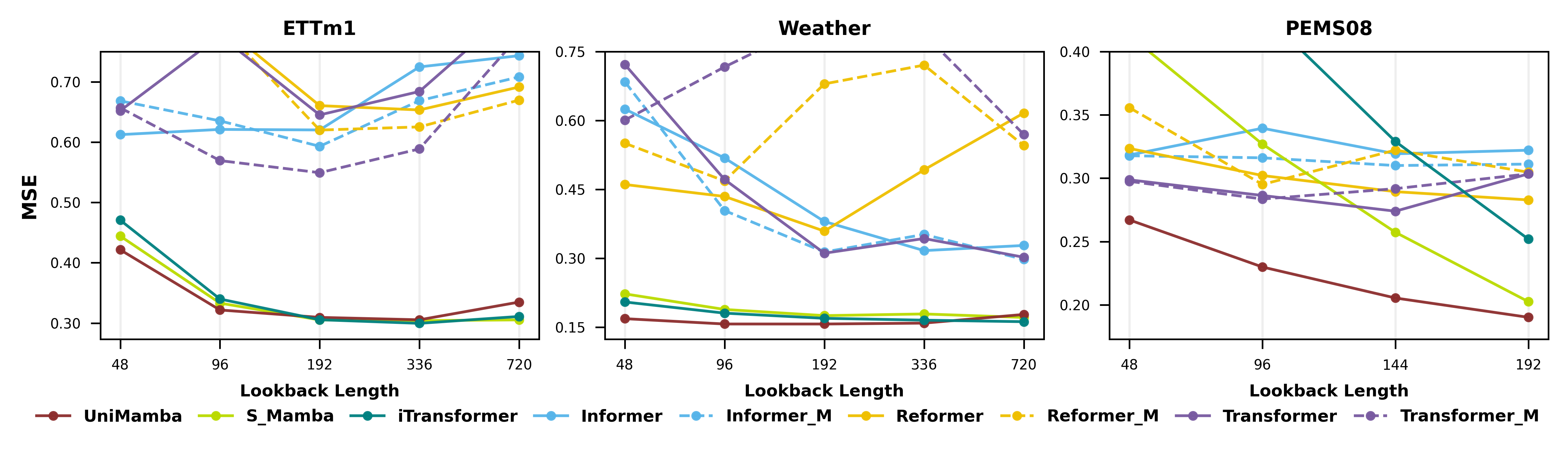}
    \caption{Prediction error values of \model\ and baseline models with increasing lookback length.
}
    \label{fig:lookback}
\end{figure*}

Figure~\ref{fig:lookback} presents the effect of varying lookback window lengths on forecasting accuracy across ETTm1, Weather, and PEMS08. It can be observed that \textbf{\model} attains the lowest MSE values under short and medium-length lookback configurations, reflecting its remarkable ability to adapt to both short-term and long-horizon temporal dependencies. In contrast to Transformer-based baselines, whose performance often exhibits substantial fluctuations or degradation as the lookback length increases, \model\ maintains steady or even enhanced predictive accuracy, underscoring its capacity for efficient long-range dependency modeling. On ETTm1 and Weather, accuracy improvements tend to converge beyond 192 lookback length, implying that trends and periodic components are effectively captured within this range. On PEMS08, \model\ continues to benefit from extended historical contexts, highlighting its pattern capture and signal reconstruction capabilities with high-dimensional series. It's worth noting that MSE increases when lookback length reaches 720, possibly caused by over-comprehensive consideration of all patterns in the lookback sequence and introducing unnecessary information. Overall, these results confirm that \model\ achieves an optimal utilization of given contextual series by effectively extracting and analyzing temporal patterns inside.

\begin{figure}[!t]
    \centering
    \includegraphics[width=\linewidth]{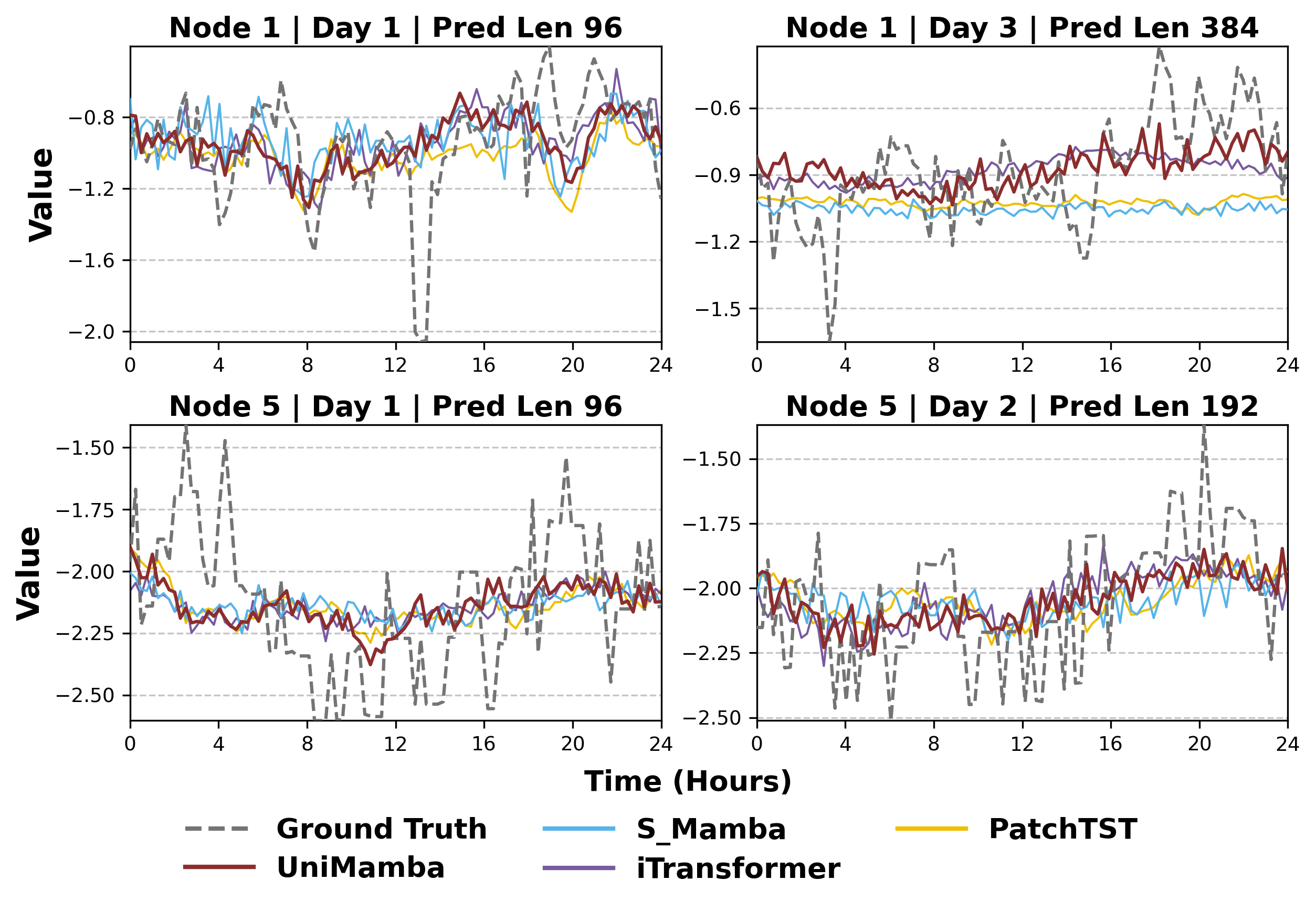}
    \caption{Case Study on ETTm2}
    \label{fig:case-study-square}
\end{figure}

\subsection{Case Study}
ETTm2 is a dataset of electricity transformer temperatures. It contains periodic patterns of transformer and oil temperatures in varied time cycles, reflecting real industrial conditions. Figure~\ref{fig:case-study-square} presents prediction comparisons among \model, S-Mamba, iTransformer, PatchTST and ground truth. Subfigures represent temperature curves of HUFL and LUFL nodes across days, generated with varied predicted lengths. It shows that \model\ closely follows the ground truth, maintaining smooth and accurate trends even for long prediction horizons. Competing models exhibit noticeable deviations and misjudgment of trends, particularly in regions with rapid fluctuations. The strong alignment of \model\ with ground truth shows its capability of accurate prediction and mutation identification. This stability attributes to functionalities provided by different components in the unified framework. Overall, \model\ delivers faithful and reliable reconstruction of temporal patterns, demonstrating its potentiality and generality in real industrial production environments.

\section{Conclusion}
This paper presents \textbf{\model}, a unified spatial temporal forecasting framework that combines \xs{frequency-domain analysis and reconstruction, temporal convolution, state space modeling, and attention-based fusion in a single architecture. Through the integration of FFT and Laplace reconstruction, reverse Mamba, TCN, and Spatial Temporal Attention mechanism,} \model\ can effectively capture variate and temporal dependencies and forecast time series accurately. Experimental results show that \model\ matches or surpasses best forecasting models in accuracy and training efficiency, maintaining stability even under noise and varying input conditions. Its harmonious modular design \xs{enables deployment in large-scale or real-time time series forecasting scenarios.}

In future work, we plan to extend \model\ to handle irregularly sampled signals and multimodal temporal data while exploring adaptive mechanisms for continual training. These enhancements aim to further strengthen \model ’s practicality and generalization in complex, dynamic spatial temporal environments.

\bibliographystyle{IEEEtran}
\bibliography{IEEEabrv,ref}

\end{document}